# A two-stage architecture for stock price forecasting by combining SOM and fuzzy-SVM

[1]Duc-Hien Nguyen, [2]Manh-Thanh Le
Hue University
Hue, VietNam
[1]hiencit@gmail.com
[2]lmthanh@hueuni.edu.vn

*Abstract*— **This paper proposed a model to predict the stock price based on combining Self-Organizing Map (SOM) and fuzzy – Support Vector Machines (f-SVM). Extraction of fuzzy rules from raw data based on the combining of statistical machine learning models is the base of this proposed approach. In the proposed model, SOM is used as a clustering algorithm to partition the whole input space into several disjoint regions. For each partition, a set of fuzzy rules is extracted based on a f-SVM combining model. Then fuzzy rules sets are used to predict the test data using fuzzy inference algorithms. The performance of the proposed approach is compared with other models using four data sets.**

*Keywords- Fuzzy rules; Support vector machine - SVM; Self-Organizing Map - SOM; Stock price forecasting; Data-driven model*

I. INTRODUCTION

Nowadays, time series forecasting, especially predicting the stock market has attracted a lot of interest from many scientists. For the ultimate objective of increasing the accuracy of predicting results, many researchers have made contributions to conducting and improving various models and solutions. The current stock market prediction is approached in two methods, either stock price prediction or the trend of stock price past n-days.

Today, the application of data mining and statistical machine learning techniques are two common approaches used to predict stock market movements. Many researches in [7], [8], [14], [16], [17] proposed applications of Artificial Neutral Network, Support Vector Machine - SVM, Hidden Markov Model - HMM in stock market prediction. In order to make more effective and accurate predictions, various combined models with different forecasting methods [4], [9], [11] are researched and proposed by researchers. A model based on the Fuzzy model combined with Support Vector Machine is a new trend of research, called data-driven model [5], [6], [10], whose purpose is to extract fuzzy rules from Support Vector Machine as basic functions for fuzzy system. One of the challenges for the data-driven model is automatic learning from data whose size is large but representativeness is limited. In addition, avoidance of the explosion in the number of fuzzy-rules is also a problem which needs solving.

In order to resolve the large sizes of the data set problem in data-driven model, combination of a data clustering algorithm such as k-Means, SOM,…is a new approach which used to divide the large sizes of the data in to several smaller sizes of data set [4], [11]. The main purpose of this study is to deal with the large size of the data, minimize the quantity, simplify fuzzy rules from the data; we propose a model combining SOM and SVM for fuzzy rule extraction in stock price prediction. The fuzzy rules set in small amounts will create favourable conditions for human experts to understand, dissect, evaluate, and optimize to improve the efficiency of fuzzy rules-based inference system.

The rest of this paper is organized as follows. Section 2 briefly describes the theory to support vector machines, fuzzy model and the relationships between the two models; then introduces the f-SVM method for extraction of fuzzy rules from SVMs. Section 3 presents SOM which has been widely used in data clustering. Section 4 introduces the proposed model which can produce fewer fuzzy rules based on the combination between SOM and f-SVM to predict stock market. The results obtained from the proposed model are demonstrated in comparison to other models, which will be presented in section 5. In section 6, we present the conclusion and future work.

II. FUZZY RULE EXTRACTION METHOD FROM SUPPORT VECTOR MACHINES – F-SVM ALGORITHM

Support vector machine (SVM), which is proposed by Vapnik, is a new machine learning method based on the Statistical Learning Theory and is a useful technique for data classification [2]. SVM has been recently introduced as a technique for solving regression estimation problems [5], [8], [11], and has also been used in finding fuzzy rules from numerical [5], [6], [10]. In the regression estimation task, the basic theory of SVM [2] can be briefly presented as follows:

Given a set of training data $\{(x_1, y_1), \ldots, (x_l, y_l)\} \subset X \times \mathcal{R}$, where X denotes the space of input patterns. The goal of $\varepsilon$ Support vector regression is to find a function $f(x)$ that has at most $\varepsilon$ deviation from the actually obtained targets $y_i$ for all the training data, and at the same time is as flat as possible. That is, the errors would be ignored as long as they are less than $\varepsilon$, but any deviation bigger than this would not be accepted.

$$f(x) = \sum_{i=1}^{l}(\alpha_i - \alpha_i^*) K(x_i, x) + b \qquad (1)$$

Subject to

$$\sum_{i-1}^{l} (\alpha_i - \alpha_i^*) = 0, \text{ and } C \geq \alpha_i, \alpha_i^* \geq 0, \forall i, \quad (2)$$

Where, the constant C which determines the trade-off of error margin between the flatness of f(x) and the amount of deviation in excess of ε that is tolerated; $\alpha_i, \alpha_i^*$ are Lagrange multipliers; and $K(x_i, x)$ is a Kernel function defined as

$$K(x_i, x_j) = \langle \Phi(x_i), \Phi(x_j) \rangle \quad (3)$$

where $\Phi(x_i)$ is a nonlinear function mapping.

The input points $x_i$ with $(\alpha_i - \alpha_i^*) \neq 0$ are called support vectors (SVs).

On the other hand, fuzzy rule-base that generally consists of set of IF-THEN rules is the core of the fuzzy inference [5]. Suppose there are M fuzzy rules, it can be expressed as follows:

$R_j$: IF $x_1$ is $A_1^j$ and $x_2$ is $A_2^j$ and ... and $x_n$ is $A_n^j$ THEN y is $B^j$,

for j = 1..M $\quad (4)$

where $x_i(i = 1,2,...n)$ are the input variables; y is the output variable of the fuzzy system; and $A_i^j$ and $B^j$ are linguistic terms characterized by fuzzy membership functions $\mu_{A_i^j}(x_i)$ and $\mu_{B^j}(y)$, respectively.

The inference process is shown as below: 1) membership values activation. The membership values of input variables are computed as t-norm operator

$$\prod_{i=1}^{n} \mu_{A_i^j}(x_i) \quad (5)$$

2) the final output can be computed as

$$f(x) = \frac{\sum_{j=1}^{M} \bar{z}^j \left( \prod_{i=1}^{n} \mu_{A_i^j}(x_i) \right)}{\sum_{j=1}^{M} \prod_{i=1}^{n} \mu_{A_i^j}(x_i)} \quad (6)$$

where $\bar{z}^j$ is the output value when the membership function $\mu_{B^j}(y)$ achieves its maximum value.

In order to let equation (1) and (6) be equivalent, at first we have to let the kernel functions in (1) and the membership functions in (6) be equal. The Gaussian membership functions can be chosen as the kernel functions since the Mercer condition [15] should be satisfied. Besides, the bias term b of the expression (1) should be 0.

While the Gaussian functions are chosen as the kernel functions and membership functions, and the number of rules - M equal to the number of support vectors - l, then (1) and (6) become:

$$f(x) = \sum_{i=1}^{l} (\alpha_i - \alpha_i^*) \exp\left(-\frac{1}{2}\left(\frac{x_i - x}{\sigma_i}\right)^2\right) \quad (7)$$

and

$$f(x) = \frac{\sum_{j=1}^{l} \bar{z}^j \exp\left(-\frac{1}{2}\left(\frac{x_j - x}{\sigma_j}\right)^2\right)}{\sum_{j=1}^{l} \exp\left(-\frac{1}{2}\left(\frac{x_j - x}{\sigma_j}\right)^2\right)} \quad (8)$$

The inference of fuzzy systems can be modified as [3]

$$f(x) = \sum_{j=1}^{l} \bar{z}^j \exp\left(-\frac{1}{2}\left(\frac{x_j - x}{\sigma_j}\right)^2\right) \quad (9)$$

and the center of Gaussian membership functions are selected as

$$\bar{z}^j = (\alpha_i - \alpha_i^*) \quad (10)$$

Then, the output of fuzzy system (6) is equal to the output of SVM (1). However, the equivalence has some shortcomings: 1) the modified fuzzy model removes the normalization process; therefore, the modified fuzzy model sacrifices the generalization. 2) the interpretability cannot be provided during the modification.

An alternative approach is to set the kernel function of SVMs as

$$K(x_i, x) = \frac{\exp\left(-\frac{1}{2}\left(\frac{x_i - x}{\sigma_i}\right)^2\right)}{\sum_{i=1}^{l} \exp\left(-\frac{1}{2}\left(\frac{x_i - x}{\sigma_i}\right)^2\right)} \quad (11)$$

Consequently, the output of SVMs becomes

$$f(x) = \frac{\sum_{i=1}^{l} (\alpha_i - \alpha_i^*) \exp\left(-\frac{1}{2}\left(\frac{x_i - x}{\sigma_i}\right)^2\right)}{\sum_{i=1}^{l} \exp\left(-\frac{1}{2}\left(\frac{x_i - x}{\sigma_i}\right)^2\right)} \quad (12)$$

We only have to set the centre of membership functions to $(\alpha_i - \alpha_i^*)$, then we can assure the output fuzzy systems (12) and the output of the SVMs (7) are equal. Notably, the expression (11) can only be achieved when the number of support vectors, l, is known previously.

From the analysis of the similarity of SVMs and fuzzy systems above, we propose F-SVM algorithm in Figure 1 that allows extracting fuzzy rules from SVMs.

Parameters of membership functions can be optimized utilizing gradient decent algorithms of adaptive networks. In general, optimal fuzzy sets have different variances, while the kernel functions have the same ones. In order to obtain a set of optimal fuzzy with different variances, we can adopt methods such as gradient decent algorithms or GAs. We derive the following adaptive algorithm to update the parameters in the fuzzy membership functions:

$$\sigma_i(t+1) = \sigma_i(t)\delta\varepsilon_{1,i}\left[\frac{(x-c)^2}{\sigma^3}\exp\left(-\frac{(x-c)^2}{2\sigma^2}\right)\right] \quad (13)$$

$$c_i(t+1) = c_i(t)\delta\varepsilon_{1,i}\left[\frac{-(x-c)}{\sigma^2}\exp\left(-\frac{(x-c)^2}{2\sigma^2}\right)\right] \quad (14)$$

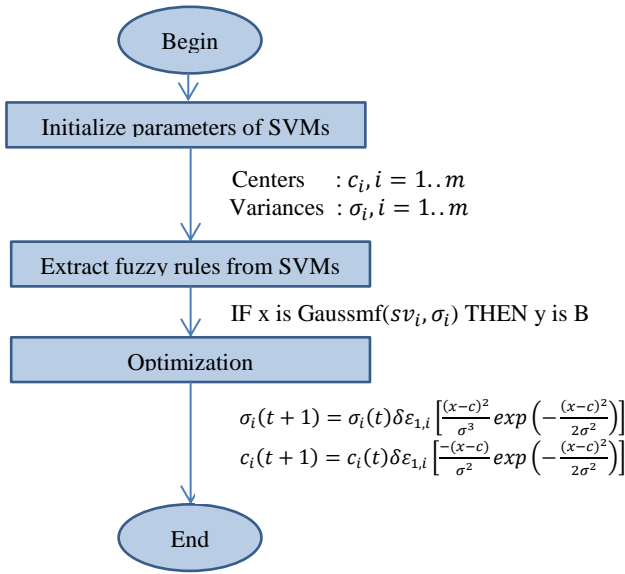

Figure 1. Block diagram of f-SVM algorithm.

## III. DATA CLUSTERING USING SELF-ORGANIZING MAPS

SOM (Self-Organizing Map) is a type of artificial neural network that is trained by using unsupervised learning that was introduced by Kohonen [12], [18]. This model was proposed as an effective solution toward the recognition and control of robots. In SOM, the output neurons are usually organized into D-dimensional map in which each output neuron is connected to each input neuron. The arrangement of neuron is a regular spacing in a hexagonal or rectangular grid. The structure of a Kohonen SOM is shown in Figure 2.

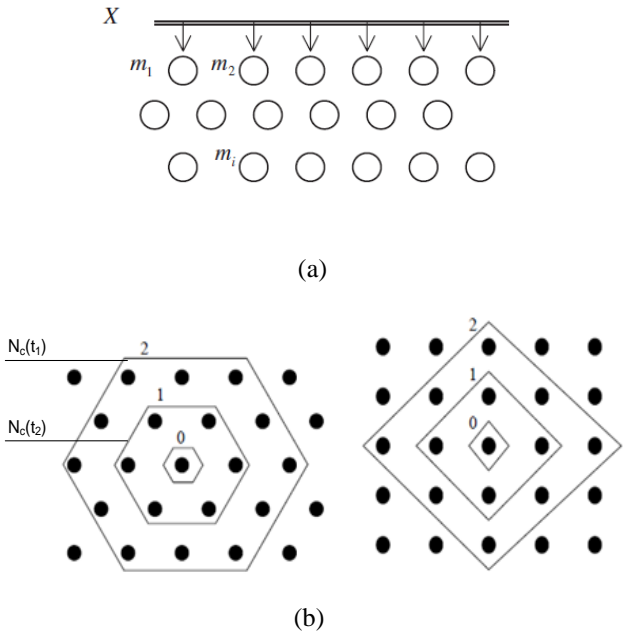

Figure 2. (a) An SOM example. (b) The distribution of rectangular and hexagonal SOM

As Figure 2, in SOM, each neuron is associated with a reference vector $m_i$ and neighborhood range $N_c$. The reference vector has to be the same size as the size of the input vector and is used as the measure of closeness between input vectors. The neighborhood range is a symmetric function and also monotonically decreases with the distance between neurons in the map and centre neuron (wining neuron).

The SOM generalizes the wining neuron to its neighbors on the map by performing the training algorithm for the input vectors. The final result is that the neurons on the map ordered: neighboring neurons have similar weight vectors (Figure 2b). SOM is widely used for clustering because after training, the output neurons of SOM are automatically organized into a meaningful two-dimensional order in which similar neurons are closer to each other than the more dissimilar ones in terms of the reference vectors, thus keep close in the output space for the data points which are similar in the input space. Recent studies have suggested using SOM as quite an effective solution for stock market data [4], [11]. In this research, we do not have desire for in-depth analyses of machine learning SOM, the details of SOM have been presented in [12], [18]. Many researches in [4], [11] have also improved the effectiveness of combining SOM and SVMs model for data clustering both from theoretical and empirical analysis.

## IV. THE STOCK PRICE FORECASTING MODEL BASED ON COMBINATION OF SOM AND F-SVM

In this research, the purpose to predict stock market and we propose a fuzzy inference model based on fuzzy rules extraction method from transaction history data. The model, which extracts fuzzy rules from data, is constructed by combining the cluster technique using SOM and f-SVM algorithm. A diagram of stock market prediction model is presented in Figure 3.

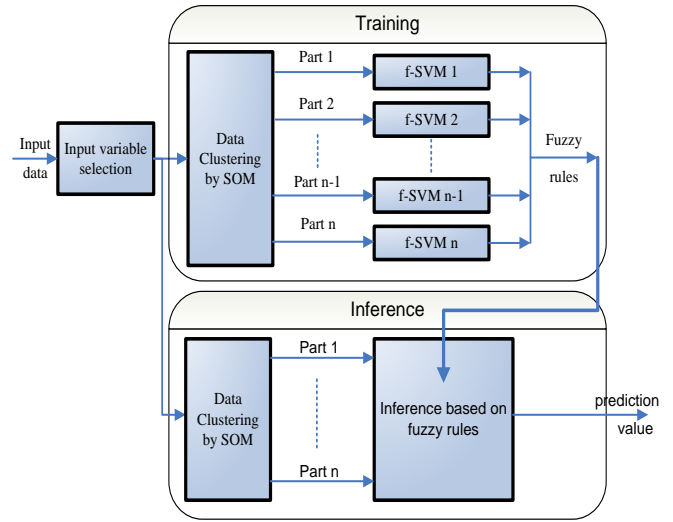

Figure 3. Block diagram of forecasting model.

### A. Input variable selection

The results of other authors on stock market predictability showed that there are many ways to select input variables, such as using daily stock market index <opening, high, low, closing price> [8], [17], macroeconomic indicators[1], ,… In this model, we have chosen stock market index as input variable.

According to the analysis and evaluation of L.J. Cao and Francis E.H. Tay in [8], 5-day relative difference in percentage of price - RDP is more effective, especially in the stock market prediction performance. In this model, we select the input variables based on the proposal and calculation of L.J. Cao and Francis E.H. Table 1 presents selected variables and their calculations.

TABLE I. TABLE TYPE STYLES

| Symbol | Variable | Calculation |
|---|---|---|
| $x_1$ | EMA100 | $P_i - \overline{EMA_{100}}(i)$ |
| $x_2$ | RDP-5 | $(P(i) - P(i-5))/P(i-5) * 100$ |
| $x_3$ | RDP-10 | $(P(i) - P(i-10))/P(i-10) * 100$ |
| $x_4$ | RDP-15 | $(P(i) - P(i-15))/P(i-15) * 100$ |
| $x_5$ | RDP-20 | $(P(i) - P(i-20))/P(i-20) * 100$ |
| $y$ | RDP+5 | $(\overline{P(i+5)} - \overline{P(i)})/\overline{P(i)} * 100$ <br> $\overline{P(i)} = \overline{EMA_3}(i)$ |

where P(i) is closing price of the *i-th* day, and EMA$_m$(i) is *m-day* exponential moving average closing price of the *i-th* day.

### B. Clustering data by SOM

For data mining application, typically we work with a large volume data while many algorithms are ineffective for large data set. A common approach to solve this problem is split input data into smaller clusters, then apply the learning algorithm to each cluster and synthesize the results of simulation studies [13]. Moreover, one of the problems in financial time series forecasting is that time series are non-stationary. Statistics of stock prices depend on different factors such as economic growth and recession, political situation, environment, calamity… There is a limitation to find out stock price prediction rules based on historical market data. In the proposed model, the SOM is used to decompose the whole input space into regions where data points with similar statistical distributions are grouped together, so as to capture the non-stationary property of financial series.

The results of clustering of data by SOM provide an effective way to solve the two problems [4]: 1) Reducing data to a small number of dimensions is useful for increasing the speed of the model. 2) The data clusters are equivalent in statistical distributions to avoid interference.

### C. Fuzzy rules extraction by f-SVM

Each cluster which was clustered by SOM will be trained for respective f-SVM machine to extract fuzzy rules. As shown in detail in the section 2, f-SVM machine extracts the fuzzy rules from each cluster of input data based on support vectors obtained from the SVM module which is integrated inside. By extracting fuzzy rules from SVM we will obtain rule sets in form:

IF $x_1$ is $Gaussmf(sv_1^i, \sigma_1^i)$ and $x_2$ is $Gaussmf(sv_2^i, \sigma_2^i)$

and ... $x_j$ is $Gaussmf(sv_j^i, \sigma_j^i)$ ... THEN $y$ is $B^i$

where $Gaussmf(sv_j^i, \sigma_j^i)$ is Gauss membership function.

### D. Stock market prediction based on fuzzy rules

Extraction of fuzzy rules from f-SVM machine is an effective method for predicting stock market movements. Clustering low size data will reduce the complexity of fuzzy inference algorithm.

The above fuzzy rules in data mining have a certain distance to the understanding of human experts; however, fuzzy clustering is a condition for human expert to understand and evaluate these rules.

## V. EXPERIMENT AND RESULTS

In order to evaluate the performance of the proposed model, we build a test system based on Matlab Toolkit. In this system, the SOM tool for Matlab is used to partition input data into several buckets, that toolbox developed by Esa Alhoniemi, Johan Himberg, Juha Parhankangas and Juha Vesanto [20]. The SOM Toolbox can be downloaded from http://www.cis.hut.fi/projects/somtoolbox/. To produce support vectors from training data we used LIBSVM – a library for Support vector Machines developed by Chih-Chung Chang and Chih-Jen Lin [19], which can be downloaded from http://www.csie.ntu.edu.tw/~cjlin/libsvm/. Finally, we use AVALFIS function in Matlab Fuzzy Logic Toolkit to infer stock market prediction based on producing fuzzy rules.

### A. Data sets

The experimental data source was chosen from famous individual companies and composite indexes in America includes IBM Corporation stock (IBM), the Apple inc. stock (APPL), the Standard & Poor's stock index (S&P500), and the Down Jones Industrial Average index (DJI). All data used in this work are downloaded from Yahoo Finance http://finance.yahoo.com/

The daily data including Close-Price of four stocks are used as data sets for experimental. List of data sources are presented in Table 2. For each data set, the data is divided into two subsets according to the time sequence - training and testing subsets. With the objective of maximizing the size of training data to increase the coverage capability of training data samples, there are only 200 data samples used for the testing subset and all of the rest data are used for the training subset.

TABLE II. THE DATA SOURCE INFORMATION

| Stock name | Time period | Training data | Testing data |
|---|---|---|---|
| IBM Corporation stock (IBM) | 03/01/2000 - 30/06/2010 | 2409 | 200 |
| Apple inc. stock (APPL) | 03/01/2000 - 30/06/2010 | 2409 | 200 |
| Standard & Poor's stock index (S&P500) | 03/01/2000 - 23/12/2008 | 2028 | 200 |
| Down Jones Industrial Average index (DJI) | 02/01/1991 - 28/03/2002 | 2352 | 200 |

### B. Performance metrics

The performance metrics used to evaluate in this study are the normalized mean squared error (NMSE), mean absolute error (MAE), and directional symmetry (DS) [4][8][11]. Among them, NMSE and MAE are measures of deviation between the actual value and the forecast value. DS provides an indication of the predicted direction of RDP+5 given in the form of percentages. The predicted results are better if the values of NMSE and MAE are small, while large value of DS is better. The definitions of these metrics can be found in Table 3.

TABLE III. METRICS

| Metrics | Calculation |
|---|---|
| NMSE | $\frac{1}{\sigma^2 n}\sum_{i=1}^{n}(y_i - \hat{y}_i)^2$ <br> $\sigma^2 = \frac{1}{n-1}\sum_{i=1}^{n}(y_i - \bar{y})^2$ <br> $\bar{y} = \sum_{i=1}^{n} y_i$ |
| MAE | $\frac{1}{n}\sum_{i=1}^{n}\left|y_i - \hat{y}_i\right|$ |
| DS | $\frac{100}{n}\sum_{i=1}^{n} d_i$ <br> $d_i = \begin{cases} 1 & (y_i - y_{i-1})(\hat{y}_i - \hat{y}_{i-1}) \geq 0 \\ 0 & \text{otherwise} \end{cases}$ |

n is the total number of data patterns
y and ŷ represent the actual and predicted output value

*C. Experimental Results*

Table 4 presents a group of fuzzy rules are produced from S&P500 stock data.

TABLE IV. A GROUP OF FUZZY RULES ARE PRODUCED FROM S&P500 STOCK DATA

| Rule | Detail |
|---|---|
| $R_1$ | IF $x_1$=Gaussmf(0.09,-0.11) and $x_2$=Gaussmf(0.09,-0.12) and $x_3$=Gaussmf(0.09,-0.04) and $x_4$=Gaussmf(0.09,-0.10) and $x_5$=Gaussmf(0.09,-0.09) THEN y=0.10 |
| $R_2$ | IF $x_1$=Gaussmf(0.10,-0.01) and $x_2$=Gaussmf(0.09,-0.06) and $x_3$=Gaussmf(0.10,0.04) and $x_4$=Gaussmf(0.10,-0.10) and $x_5$=Gaussmf(0.10,-0.12) THEN y=0.57 |
| $R_3$ | IF $x_1$=Gaussmf(0.09,0.02) and $x_2$=Gaussmf(0.10,0.02) and $x_3$=Gaussmf(0.09,0.08) and $x_4$=Gaussmf(0.10,-0.08) and $x_5$=Gaussmf(0.10,-0.13) THEN y=-0.02 |
| $R_4$ | IF $x_1$=Gaussmf(0.10,-0.04) and $x_2$=Gaussmf(0.10,-0.08) and $x_3$=Gaussmf(0.10,0.02) and $x_4$=Gaussmf(0.09,-0.08) and $x_5$=Gaussmf(0.09,-0.11) THEN y=-0.29 |
| R5 | IF x1=Gaussmf(0.10,-0.03) and x2=Gaussmf(0.09,-0.06) and x3=Gaussmf(0.10,0.03) and x4=Gaussmf(0.09,-0.10) and x5=Gaussmf(0.09,-0.13) THEN y=-0.38 |

We conduct an experiment to compare the results between the proposed model which predicts stock market based on fuzzy rules extraction and other models such as the RBN model and the hybrid model of SOM and SVM with the same testing data (200 samples). RBN model was built on a generalized regression neural network which is a type of Radial Basis Network (RBN). The generalized regression neural network is often used for prediction problems in [7], [14], [16]. The hybrid model of SOM and SVM was proposed to improving the effectiveness of time-series forecasting, especially stock market prediction [4], [11]. Moreover, we have compared with the experiment results of ANFIS model (Adaptive Neural Fuzzy Inference System). ANFIS model is a fuzzy neural network model which was proposed and standardized in Matlab. ANFIS has been applied in several studies in prediction problems. The prediction performance is evaluated using the following statistical metrics: NMSE, MAE, and DS. The results of the proposed model and other models are shown in Table 5&6.

TABLE V. RESULTS OF RBN AND SOM+ANFIS

| Stock code | RBN | | | SOM+ANFIS | | |
|---|---|---|---|---|---|---|
| | *NMSE* | *MAE* | *DS* | *NMSE* | *MAE* | *DS* |
| IBM | 1.1510 | 0.0577 | 43.72 | 1.2203 | 0.0617 | 47.74 |
| APPL | 1.3180 | 0.0475 | 45.73 | 2.8274 | 0.0650 | 49.75 |
| SP500 | 1.2578 | 0.1322 | 51.76 | 1.7836 | 0.1421 | 48.24 |
| DJI | 1.0725 | 0.1191 | 50.75 | 1.7602 | 0.1614 | 49.75 |

TABLE VI. RESULTS OF SOM+SVM AND SOM+F-SVM

| Stock code | SOM+SVM | | | SOM+f-SVM | | |
|---|---|---|---|---|---|---|
| | *NMSE* | *MAE* | *DS* | *NMSE* | *MAE* | *DS* |
| IBM | 1.1028 | 0.0577 | 44.22 | 1.0324 | 0.0554 | 50.75 |
| APPL | 1.1100 | 0.0445 | 52.76 | 1.0467 | 0.0435 | 53.27 |
| SP500 | 1.1081 | 0.1217 | 52.76 | 1.0836 | 0.1207 | 53.27 |
| DJI | 1.0676 | 0.1186 | 50.25 | 1.0459 | 0.1181 | 51.76 |

The experiment results in Table 5&6 demonstrates that the MNSE and MAE of SOM+f-SVM model are smaller than RBN and SOM+ANFIS, indicating that there is a smaller deviation between the actual and predict values in SOM+f-SVM. Moreover, the DS (Directional Symmetry) of the proposed model is higher than RBN and SOM+ANFIS. This shows that the predictions of SOM+f-SVM are more accurate than those of two other models. The comparison between the SOM+f-SVM model and SOM+SVM model (L.J. Cao and Francis E.H in [4]) is shown in Table 5, which shows that the values of MNSE, MAE and DS of the proposed model have not improvement significantly. This is obvious, because f-SVM algorithm used in proposed model perform extracts fuzzy rules from SVMs. The SOM+SVM model is used as "black-box" learning and inference processes. Otherwise, the proposed model allows producing a set of fuzzy rules and the inference processes will be performed using these rules. Results of learning process which is fuzzy rules extraction from SVMs have gradually clarified "black-box" model of SVMs. Based on the set of extracted rules, the human experts can understand and interact to improve the efficiency of using set of rules for inference process. In addition, using SOM for data clustering to split the input data into several smaller datasets brings the following effects: reducing the size of input data and thereby reducing the complexity of the algorithm, the generated rules will be split into several clusters, respectively. It helps human experts to understand and analysis fuzzy rules easily.

VI. CONCLUSIONS

In this study, we proposed an F-SVM algorithm to extract fuzzy rules from SVM; then we developed a stock market prediction model based on combination SOM and F-SVM. The experiment results showed that the proposed model has been used to predict stock market more effectively than the previous models, reflected in better values of three parameters: NMSE, MAE and DS. In addition, data cluster with SOM has been used to improve execution time of algorithms significantly in this model. Otherwise, as shown in section 5.2 of this paper, the efficacy of the proposed model is the use of extraction of fuzzy rules which is a form of splitting set of rules; it helped in analyzing fuzzy rules easily. However, there are some drawbacks in SVM model: if it improves the accuracy of the model, the number of SVs will be increased; which causes an increase of the number of fuzzy rules. Thus, the system is more complex, especially the interpretability of the set of rules will decrease and then, human experts have difficulties in understanding and analyzing those rule sets.

In future work, we will concentrate on finding solutions to improve the interpretability of the sets of rules which are extracted from SVMs. After solving this problem, we gain the basis for analyzing the sets of rules and then optimize them in order to improve the effect of prediction.